\documentclass{article}

\usepackage[preprint]{neurips_2025}

\usepackage[utf8]{inputenc} % allow utf-8 input
\usepackage[T1]{fontenc}    % use 8-bit T1 fonts
\usepackage[svgnames,dvipsnames,table]{xcolor}
\usepackage[colorlinks=true, pdfusetitle, urlcolor=RoyalBlue, linkcolor=RoyalBlue, citecolor=RoyalBlue]{hyperref}       % hyperlinks
\usepackage{url}            % simple URL typesetting
\usepackage{booktabs}       % professional-quality tables
\usepackage{amsfonts}       % blackboard math symbols
\usepackage{nicefrac}       % compact symbols for 1/2, etc.
\usepackage{microtype}      % microtypography
\usepackage{xcolor}         % colors
\usepackage{cleveref}

\usepackage{enumitem}
\usepackage{graphicx}
\usepackage{wrapfig}
\usepackage{kantlipsum}
\usepackage{amssymb}
\usepackage{pifont}% http://ctan.org/pkg/pifont
\newcommand{\xmark}{\ding{55}}

\title{From Calibration to Collaboration: LLM Uncertainty Quantification Should Be More Human-Centered}

\newcommand{\USC}{University of Southern California}

\author{%
    Siddartha Devic \quad Tejas Srinivasan \\
    \textbf{Jesse Thomason} \quad \textbf{Willie Neiswanger} \quad \textbf{Vatsal Sharan} \\
    \USC \\
    \texttt{\{devic, tejas.srinivasan\}@usc.edu}
}

\begin{document}

\maketitle

\begin{abstract}
    Large Language Models (LLMs) are increasingly assisting users in the real world, yet their reliability remains a concern. 
    Uncertainty quantification (UQ) has been heralded as a tool to enhance human-LLM collaboration by enabling users to know when to trust LLM predictions. 
    We argue that \textbf{current practices for uncertainty quantification in LLMs are not optimal for developing \emph{useful} UQ for human users making decisions in real-world tasks}. 
    Through an analysis of 40 LLM UQ methods, we identify three prevalent practices hindering the community's progress toward its goal of benefiting downstream users: 1) evaluating on benchmarks with low ecological validity; 2) considering only epistemic uncertainty; and 3) optimizing metrics that are not necessarily indicative of downstream utility. 
    For each issue, we propose concrete user-centric practices and research directions that LLM UQ researchers should consider. Instead of hill-climbing on unrepresentative tasks using imperfect metrics,
    we argue that the community should adopt a more human-centered approach to LLM uncertainty quantification.
\end{abstract}

\section{Introduction}

Classical machine learning and forecasting communities have long been focused on the tantalizing possibility of predicting outcomes and future events.
A key, highly related goal is to accurately estimate when model predictions can safely be trusted or otherwise ignored by a human user \citep{guo2017calibration, valdenegro2024dilemma, marusich2024using}.
This goal forms the basis for methods that estimate predictive uncertainty of machine learning models, which we will collectively refer to as \emph{uncertainty quantification} (UQ) methods~\citep{sullivan2015introduction}.

While UQ has traditionally been studied in classical machine learning models, the advent of large language models (LLMs) has precipitated a host of research creating and adapting UQ methods suited to this new paradigm. 
The primary reason for quantifying LLM uncertainty is \emph{user-centric}: the motivation behind many proposed LLM UQ methods is to enable users to know when to trust LLM outputs~\citep{xiong2023can, lin2024generating, kuhn2023semantic, duan2023shifting}, improving the reliability and trustworthiness \citep{duan2023shifting} of LLMs and allowing for their proliferation into high-stakes domains \citep{bakman2024mars}. 
LLM UQ methods include calibrating confidence scores displayed to users for particular generations or claims within generations \citep{detommaso2024multicalibration, yang2024verbalized}, altering the language used in generations to better represent the uncertainty that the model has in its generations \citep{band2024linguistic, stengel2024lacie}, and generating predictions sets with coverage guarantees in a conformal prediction framework \citep{quach2024conformal}.

Although LLM UQ methods have demonstrated the ability to quantitatively improve certain LLM \emph{capabilities}---for example, via post-processing \citep{li2024graph} or sampling and aggregating multiple generations \citep{kuhn2023semantic}---we argue that the original motivation to foster better reliance by users has fallen by the wayside. 
In particular, human studies are few and far between in the LLM UQ literature, and consequently we lack a comprehensive understanding of how useful LLM UQ methods are in assisting humans on real-world decision making tasks, or when these methods increase the transparency of recommendations and decisions generated by an LLM. 
Instead, many proposed LLM UQ methods are focused on hill-climbing flawed calibration metrics capturing only epistemic uncertainty on a narrow range of benchmarks which lack \emph{ecological validity} \citep{de2020towards}.

Currently, LLM UQ is primarily a \emph{quantitative} area, with new technical tools achieving improved benchmark scores (e.g. lower calibration errors) every month.
\textbf{In this position paper, we question the assumption that current practices for benchmarking LLM UQ methods moves us efficiently towards optimal human-AI collaboration.} 
We identify various common practices in the community that hinder us from properly evaluating the utility of LLM UQ methods for LLM-assisted decision making. 
We then lay out recommendations which we believe will push LLM UQ towards providing value for real-world users, and potentially allow for its incorporation into the many LLM-based products with which millions of humans are now interacting.

\subsection{Position and Main Claims}
We argue that the NLP community's \textbf{prevailing practices for evaluating LLM UQ methods are insufficient to benefit human users in real-world settings.}
From this user-centric viewpoint, we identify three major barriers towards widespread adoption of LLM UQ methods:
\begin{enumerate}[label=\arabic*., leftmargin=*]
    \item In Section \ref{sec:ecology}, we argue that LLM UQ methods are \textbf{primarily evaluated on benchmarks with low \emph{ecological validity}} which are not representative of real-world decision making scenarios.

    \item In Section \ref{sec:source}, we argue that benchmarks commonly used for LLM UQ evaluation \textbf{do not consider aleatoric uncertainty}, and to an extent, distributional uncertainty. These are important types of uncertainty that a user may encounter when using a large language model (LLM) as a decision aid.

    \item In Section \ref{sec:hci}, we first argue that the \emph{interaction} between LLM UQ methods and real-world users is heavily understudied and not well understood. Next, we claim that LLM UQ methods are evaluated using \textbf{metrics that have little relation with downstream utility to users}. Finally, possible \emph{uncertainty presentation schemes} to LLM users on language tasks itself is understudied.
\end{enumerate}
For each issue, we highlight concrete remedies and research directions that LLM UQ methods researchers may find illuminating.
Together, we hope that these pointed shortcomings of existing UQ practices can guide the field to providing useful algorithmic advice in real-world human-AI collaboration. We also discuss some alternative views in Section~\ref{sec:alternative-views}.

\section{Preliminaries and Methodology}

The goal of uncertainty quantification (UQ) methods in both classical machine learning and LLMs is to extract an estimate across all sources of uncertainty in the system.
UQ methods typically attempt to quantify three types of uncertainty: \emph{epistemic}, \emph{aleatoric}, and \emph{distributional} \citep{hullermeier2021aleatoric, ben2010theory}.

Epistemic uncertainty, also known as \emph{model} uncertainty, represents uncertainty about an outcome due to the limitations of the model.
Epistemic uncertainty could  arise from compute, data, or optimization limitations during training, or simply from choosing the wrong model for a particular task. 

Aleatoric uncertainty, on the other hand, represents uncertainty due to inherent randomness or noise in the \emph{data}.
The canonical example of aleatoric uncertainty is diagnosing patients with the flu: given two patients with identical symptoms, one may have the flu and the other may have allergies. 
The presence of aleatoric uncertainty supposes that outcomes assigned by nature are, to an extent, inherently random, and cannot be reduced by collecting additional training data. 
% Common causes of aleatoric uncertainty include noisy or ambiguous observations, overlapping classes, inherent randomness, or other factors that are not entirely predictable.

% Finally, in \Cref{sec:distribution-shift}, we also discuss \emph{distributional} uncertainty or \emph{distribution shift}.
Distributional uncertainty represents the common scenario when the test and train distributions are different in some quantitative aspect.
For example, a pre-trained LLM selected over other options because it achieved the highest evaluation scores on benchmarks such as MMLU or TriviaQA \citep{mmlu, triviaQA} may face a distribution shift in the presence of real-world user queries at deployment time.
A more comprehensive overview of the types of uncertainty is deferred to \citet{he2023survey, smith2024uncertainty, shorinwa2024survey}.

Uncertainty quantification methods usually attempt to estimate some subset or all of these sources of uncertainty.
LLM UQ methods in particular fall into a few broad categories.
First, there are methods which are \emph{unsupervised}: some of these methods sample multiple generations in order to test \emph{consistency} of the response, and measure uncertainty via variants of entropy \citep{xiong2023can, lin2024generating, kuhn2023semantic, duan2023shifting, malinin2021uncertainty}, and other methods simply prompt the LLM in different ways in order to probe its confidence in the answer \citep{tian2023just, kadavath2022language, yang2024verbalized}.
There are also \emph{supervised} methods which attempt to first elicit confidences via token level probabilities or special prompts, and then calibrate them in a post-hoc manner \citep{li2024graph, ulmer2024calibrating, chen2024reconfidencing, detommaso2024multicalibration, mielke2022reducing}.
Finally, there are methods which modify the LLM generation itself using tools like conformal prediction in order to better reflect the confidence of the LLM \citep{quach2024conformal, jiang2025conformal, cherian2024large, wang2025sconu}.
For a full discussion of related work, we defer to more comprehensive surveys~\citep{shorinwa2024survey,huang2024survey,liu2025uncertainty}.

\textbf{Evaluation Methodology.} 
We analyze 40 LLM UQ methods papers and their associated evaluation benchmarks to provide evidence for our claims.
We collect methods papers and the references therein from two recent LLM UQ surveys \citep{shorinwa2024survey, huang2024survey}.
In \Cref{sec:methods-annotated}, we provide a brief annotation of each selected paper, focusing on aspects relevant to our critiques such as ``Does the paper test the robustness of the method towards distributional uncertainty?'' and ``What are the benchmarks that the paper uses to evaluate its LLM UQ method?''.

\textbf{Scope and Limitations.} In this work, we only consider LLM UQ methods whose stated motivation is to \emph{improve user decision making by conveying uncertainty}.
We therefore do not consider UQ methods which primarily aim to improve capabilities of language models on reasoning, mathematical, or science benchmarks.\footnote{For example, improving GRPO-style LLM reasoning chains with UQ \citep{ye2025uncertainty} is out of scope.}
Furthermore, the scope of our claims does not extend to LLM UQ methods which may allow LLMs to act as better, fully autonomous agents in the wild \citep{liu2024dellma}.
% by, for example, making reasoning chains more efficient are excluded from discussion.
% Such capabilities research may indeed stand to benefit from robust UQ methods: instead, our observations, findings, and recommendations are intended for UQ methods which ultimately seek to improve user reliance, enable users to make safer or quantitatively better decisions, or improve the efficiency of human-AI teams.
For additional discussion on limitations of our analysis, we refer to \Cref{sec:alternative-views}.

\section{Ecological Validity of UQ Benchmarks}\label{sec:ecology}

We argue that the LLM UQ methods primarily evaluate on benchmarks with \emph{low ecological validity}~\citep{bronfenbrenner1977toward} and findings on such benchmarks are unlikely to generalize to queries posed by users in real-world settings. 
If we want LLM UQ methods to benefit users with decision making problems, then it is necessary to evaluate them on tasks that reflect real-world usage patterns. 

Ecological validity \citep{bronfenbrenner1977toward} refers to the degree to which experimental results generalize to real-world settings. 
As LLMs are being deployed to assist people in the wild, the problem of ecological validity of NLP datasets has gained focus~\citep{de2020towards}, with several efforts being made to develop benchmarks that reflect realistic user-AI interactions~\citep{wildbench, wildchat}. 
\citet{de2020towards} identify several ways in which NLP datasets lack ecological validity: synthetic language, artificial tasks, not working with prospective users, scripted interactions, and limiting to single-turn interactions. 
In this section, we assess the ecological validity of the 22 distinct benchmarks evaluated by the 40 LLM UQ methods we surveyed.\footnote{We only consider benchmarks that were used by at least two methods papers.}

\subsection{LLM UQ benchmarks represent a narrow range of tasks}

\begin{wrapfigure}{R}{0.35\textwidth}
    % \centering
    \vspace{-1.5em}
    \includegraphics[width=\linewidth]{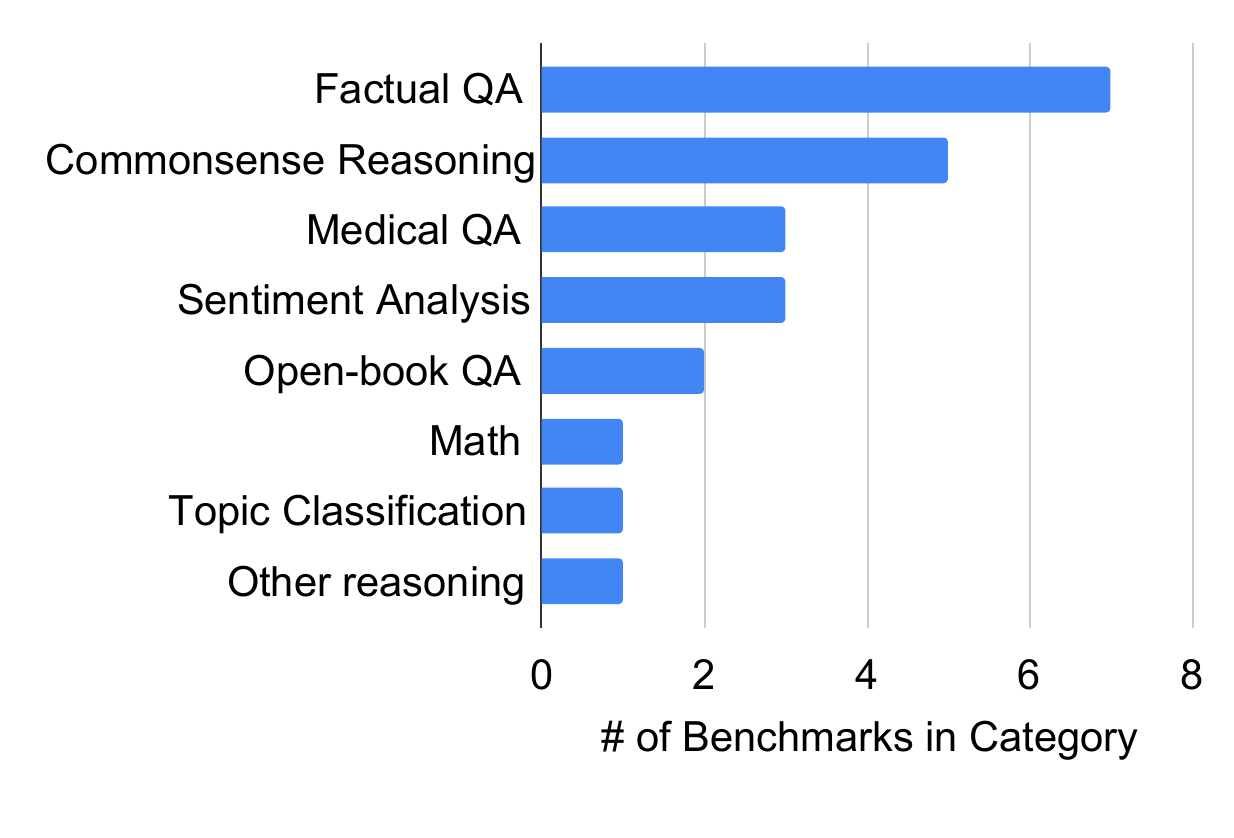}
    \vspace{-3em}
    \caption{Category-wise distribution of LLM UQ benchmarks.
    }
    \label{fig:interface}
    \vspace{-1em}
\end{wrapfigure}
Our first observation, illustrated in \Cref{fig:interface}, is that the majority of LLM UQ methods are evaluated on only factual question answering or commonsense reasoning tasks.
In particular, seventeen of the benchmarks are either question-answering (QA) or common sense reasoning tasks.
Real-world user queries are much more diverse; for instance, WildChat~\citep{wildchat} find that just 6.3\% of their LLM queries are factual question answering. 
If a majority of LLM UQ method evaluations are on only these two domains, we will not know much about evaluated methods' ability to provide useful uncertainty estimates on the wider range of decision making applications for which people use LLMs.  

\subsection{LLM UQ benchmarks do not reflect user-grounded applications}
\label{sec:benchmark-criteria}
% \usepackage{booktabs}
% \usepackage[normalem]{ulem}
\begin{table}[t]
\small
\centering
\begin{tabular}{@{}l p{2.5cm} p{2cm} p{2cm} p{2cm} p{1.7cm}@{}}
\toprule
Benchmark         & Category              & C1: The benchmark has a meaningful connection to a real-world task & C2: The benchmark inputs are representative of real-world task inputs & C3: People find the benchmark to be challenging or require notable effort & C4: Undesirable events may occur due to bad decisions. \\ \midrule
TriviaQA          & Factual QA            & \textcolor{red}{\textbf{\xmark}}                    & -                                                                                  & \textcolor{ForestGreen}{\textbf{\checkmark}}                                                                    & \textcolor{red}{\textbf{\xmark}}                    \\
Natural Questions & Factual QA            & \textcolor{ForestGreen}{\textbf{\checkmark}}                & \textcolor{ForestGreen}{\textbf{\checkmark}}                                & \textcolor{ForestGreen}{\textbf{\checkmark}}                                                                    & \textcolor{red}{\textbf{\xmark}}                    \\
MMLU              & Factual QA            & \textcolor{red}{\textbf{\xmark}}                    & -                                                                                  & \textcolor{ForestGreen}{\textbf{\checkmark}}                                                                    & \textcolor{red}{\textbf{\xmark}}                    \\
CoQA              & Open-book QA          & \textcolor{red}{\textbf{\xmark}}                    & -                                                                                  & \textcolor{ForestGreen}{\textbf{\checkmark}}                                                                    & \textcolor{red}{\textbf{\xmark}}                    \\
TruthfulQA        & Factual QA            & \textcolor{red}{\textbf{\xmark}}                    & -                                                                                  & \textcolor{red}{\textbf{\xmark}}                                                                        & -                                                                  \\
SciQ              & Factual QA            & \textcolor{red}{\textbf{\xmark}}                    & -                                                                                  & \textcolor{ForestGreen}{\textbf{\checkmark}}                                                                    & \textcolor{red}{\textbf{\xmark}}                    \\
GSM8K             & Math                  & \textcolor{red}{\textbf{\xmark}}                    & -                                                                                  & \textcolor{red}{\textbf{\xmark}}                                                                        & \textcolor{red}{\textbf{\xmark}}                    \\
BIGBench          & Other reasoning       & \textcolor{red}{\textbf{\xmark}}                    & - & \textcolor{ForestGreen}{\textbf{\checkmark}}                                                                    & \textcolor{red}{\textbf{\xmark}}                    \\
SST               & Sentiment Analysis    & \textcolor{red}{\textbf{\xmark}}                    & -                                                                                  & \textcolor{red}{\textbf{\xmark}}                                                                        & \textcolor{red}{\textbf{\xmark}}                    \\
WebQA             & Open-book QA          & \textcolor{ForestGreen}{\textbf{\checkmark}}                & \textcolor{red}{\textbf{\xmark}}                                    & \textcolor{red}{\textbf{\xmark}}                                                                        & \textcolor{red}{\textbf{\xmark}}                    \\
CommonSenseQA     & Commonsense & \textcolor{red}{\textbf{\xmark}}                    & -                                                                                  & \textcolor{red}{\textbf{\xmark}}                                                                        & \textcolor{red}{\textbf{\xmark}}                    \\
IMDB              & Sentiment Analysis    & \textcolor{ForestGreen}{\textbf{\checkmark}}                & \textcolor{ForestGreen}{\textbf{\checkmark}}                                & \textcolor{red}{\textbf{\xmark}}                                                                        & \textcolor{red}{\textbf{\xmark}}                    \\
AGNews            & Topic Classification  & \textcolor{ForestGreen}{\textbf{\checkmark}}                & \textcolor{ForestGreen}{\textbf{\checkmark}}                                & \textcolor{red}{\textbf{\xmark}}                                                                        & \textcolor{red}{\textbf{\xmark}}                    \\
Yelp Reviews      & Sentiment Analysis    & \textcolor{ForestGreen}{\textbf{\checkmark}}                & \textcolor{ForestGreen}{\textbf{\checkmark}}                                & \textcolor{red}{\textbf{\xmark}}                                                                        & \textcolor{red}{\textbf{\xmark}}                    \\
NQ-Open           & Factual QA            & \textcolor{ForestGreen}{\textbf{\checkmark}}                & \textcolor{ForestGreen}{\textbf{\checkmark}}                                & \textcolor{red}{\textbf{\xmark}}                                                                        & \textcolor{red}{\textbf{\xmark}}                    \\
AmbigQA           & Factual QA            & \textcolor{red}{\textbf{\xmark}}                    & -                                                                                  & \textcolor{ForestGreen}{\textbf{\checkmark}}                                                                    & \textcolor{red}{\textbf{\xmark}}                    \\
MNLI              & Commonsense & \textcolor{red}{\textbf{\xmark}}                    & -                                                                                  & \textcolor{red}{\textbf{\xmark}}                                                                        & \textcolor{red}{\textbf{\xmark}}                    \\
SNLI              & Commonsense & \textcolor{red}{\textbf{\xmark}}                    & -                                                                                  & \textcolor{red}{\textbf{\xmark}}                                                                        & \textcolor{red}{\textbf{\xmark}}                    \\
SWAG              & Commonsense & \textcolor{red}{\textbf{\xmark}}                    & -                                                                                  & \textcolor{red}{\textbf{\xmark}}                                                                        & \textcolor{red}{\textbf{\xmark}}                    \\
HellaSwag         & Commonsense & \textcolor{red}{\textbf{\xmark}}                    & -                                                                                  & \textcolor{red}{\textbf{\xmark}}                                                                        & \textcolor{red}{\textbf{\xmark}}                    \\
BioASQ            & MedicalQA             & \textcolor{red}{\textbf{\xmark}}                    & -                                                                                  & \textcolor{ForestGreen}{\textbf{\checkmark}}                                                                    & \textcolor{red}{\textbf{\xmark}}                    \\
MedMCQA           & MedicalQA             & \textcolor{red}{\textbf{\xmark}}                    & -                                                                                  & \textcolor{ForestGreen}{\textbf{\checkmark}}                                                                    & \textcolor{red}{\textbf{\xmark}}                    \\ \bottomrule
\end{tabular}
\vspace{1em}
\caption{An analysis of our 22 benchmarks, categorized and annotated for the criteria in Section~\ref{sec:benchmark-criteria}.}
\label{tab:benchmark_analysis}
\end{table}

In the context of human-centered explainable AI (XAI), \citet{chaleshtori2024evaluating} introduce criteria for determining the suitability of a benchmark for evaluating the human utility of model explanations. 
These criteria are not specific to explanations, and hence we adopt them for evaluating the ecological validity of LLM UQ benchmarks:
% \sdmargincomment{Should we mention LLM UQ in the statements of these criteria?}
\begin{enumerate}[label=C\arabic*., leftmargin=*]
    \item \textbf{The benchmark has a meaningful connection to a real-world task where a user might seek assistance from an LLM.}
    
    For a benchmark to be ecologically valid, it must be grounded in a real-world application. 
    Examples of benchmarks \emph{not} satisfying this criterion include CommonsenseQA~\citep{commonsenseqa}, StrategyQA~\citep{strategyqa}, and TriviaQA~\citep{triviaQA}.\footnote{We consider the benchmark's construction and input domain when deciding whether it satisfies this criterion. 
    For instance, while users may use LLM assistants to answer factual questions, TriviaQA consists of niche trivia questions. The only application in which a person might ask a trivia question without knowing its answer would be if they were participating in a contest, where they would not be allowed to use an LLM.}
    
    \item \textbf{The benchmark's inputs are representative of task inputs in real-world LLM use.}
    
    Even if the benchmark is grounded in a real-world application, its inputs must represent real inputs that a user might provide to the LLM. 
    For instance, WebQA~\citep{webqa} consists of questions that are designed to yield a specific answer known ahead of time, which is unlikely to be the case when a user is actually collaborating with an LLM.
    
    \item \textbf{The benchmark instances require notable effort from people, or people are bad at the task (when provided reasonable resources).}

    Tasks that a user can complete accurately with ease do not require them to collaborate with an LLM, rendering them unsuitable for studying LLM uncertainty quantification methods. 
    For instance, one of the most popular benchmarks for studying UQ is GSM8K~\citep{gsm8k}, which involves solving grade-school math problems which most users can solve with ease.
    
    \item \textbf{Some undesirable event may occur as a result of bad decisions.}

    Quantifying LLM uncertainty is particularly important in tasks where the user is vulnerable to some undesirable consequences as a result of ineffective human-LLM collaboration.
\end{enumerate}

\citet{chaleshtori2024evaluating} argue that only benchmarks satisfying criteria C1--C3 are suitable for studying user reliance, and among those, benchmarks also satisfying criteria C4 should be prioritized.
% For each dataset used by at least two LLM UQ method papers that we survey, we annotate these benchmarks with the four axes (C1 through C4).

In our annotation of whether LLM UQ benchmarks satisfy these criteria (Table~\ref{tab:benchmark_analysis}), we find that only 6 out of 22 benchmarks (27.3\%) satisfy C1 (grounded in a real application), and only 5 also satisfy C2 (realistic inputs). 
9 out of 22 benchmarks satisfy criterion C3 (challenging for users). 
However, only 2 benchmarks satisfy all of C1--C3: Natural Questions~\citep{naturalQuestions} and NQ-Open~\citep{nqOpen}, which consist of information seeking questions from real Google search queries. 
None of the benchmarks satisfy C4 (high-stakes), indicating that popular benchmarks for evaluating LLM UQ methods tell us little of their applicability to high-stakes decision making tasks. 
% Out full set of annotations is available in \autoref{sec:benchmark_annotations}.

\subsection{LLM UQ benchmarks reflect unrealistic task formats}
Finally, we find that 15 of the 22 benchmarks we analyzed (68.2\%) are multiple choice problems.\footnote{We consider classification tasks with a fixed set of classes, such as sentiment analysis or topic classification, to be multiple choice tasks for LLMs.} 
MCQA questions are known to have many issues, as detailed in works such as \citet{khatun2024study, llm-mcq-bias, balepur2025these}.
At a high level, the issues with MCQA evaluations include position-ordering bias, rigidity of the task format, presence of adversarially chosen distractors, brittleness to perturbations, and more.
Most importantly, real-world user queries are hardly ever multiple choice, as demonstrated in ShareGPT~\citep{ouyang2023shifted} and WildChat~\citep{wildchat}. 
Even for simple knowledge queries captured by QA benchmarks, real-world user queries are still short- or long-form generations. 
Evaluating LLM UQ methods primarily on MCQA benchmarks is not necessarily informative of their ability to estimate uncertain when the set of possible decisions is not enumerated.

\subsection{Recommendations}

We make several recommendations for selecting more appropriate evaluation benchmarks:

\begin{enumerate}[label=\textbf{R\arabic*.}, leftmargin=*]
\item Developers of LLM UQ methods should evaluate on a wider diversity of benchmarks that include applications beyond factual question answering and do not come with a fixed set of options guaranteed to contain the correct answer. 
When evaluating on MCQA benchmarks, we recommend using stable evaluation protocols as prescribed by \citet{balepur2025these}.

\item When selecting benchmarks for evaluating LLM UQ methods, we recommend evaluating benchmarks against the criterion adopted from \citet{chaleshtori2024evaluating} in \Cref{sec:benchmark-criteria}. 
Benchmarks that satisfy these criteria will provide more meaningful insights towards the applicability of LLM UQ methods in real-world human-AI collaboration.

\item We call for the community to pay more attention towards understanding the nature of uncertainty in generation tasks, such as code generation~\citep{vasconcelos2025uncertainty}, summarization~\citep{he2024can}, planning~\citep{hu2024uncertainty}, and other complex reasoning tasks that do not meet simplistic notions of ``correctness'' that factual QA tasks do.
\end{enumerate}

\section{Sources of Uncertainty Considered in LLM UQ}\label{sec:source}

In this section, we argue that many of the evaluation benchmarks used to measure the progress of LLM UQ methods fail to address two critical aspects: aleatoric uncertainty and distribution shift. 
From a user-centric viewpoint, an ideal LLM UQ method would be capable of representing \emph{all} sources of uncertainty present, not only epistemic.
However, we find a prevailing tendency in LLM UQ research to favor established QA benchmarks, despite their lack of aleatoric uncertainty.
Furthermore, despite the importance of LLM UQ methods to detect out of distribution examples \citep{lin2022teaching}, we find that nearly half of all surveyed \emph{supervised} LLM UQ methods are not evaluated on their performance under distribution shift.
% We identify two factors which contribute to insufficient evaluation on aleatoric uncertainty: (i) a limited number of benchmarks incorporate inherent aleatoric uncertainty; and (ii) a prevailing tendency in LLM UQ research to favor established QA benchmarks, even if they lack aleatoric components.
% On the other hand --- although LLM UQ methods are sometimes used to detect out of distribution examples via selective prediction or abstention --- about half of \emph{supervised} LLM UQ methods are simply not evaluated on their performance under distribution shifts.
Together, these oversights suggest that some LLM UQ methods may lack reliability in real-world scenarios containing interacting forms of uncertainty.

\subsection{Aleatoric Uncertainty}
\label{sec:aleatoric}
Human decision makers are often faced with tasks that may be inherently uncertain.
For example, the electronic patient health records available to a clinician in an emergency triage situation may not fully explain the state that the patient is in, but the clinician is still required to making a treatment decision \citep{alur2023auditing}. 
A bank may have only a few features about a potential client available --- their credit score, income, etc. --- and yet the bank needs to make a decision on whether to give out a mortgage loan or not.
In each of these cases --- due to the inherent randomness of nature, partial observability of features, or both --- true outcomes may be \emph{non-deterministic} and contain aleatoric uncertainty.
Indeed, it is often in similar high stakes situations that many envision machine assistance to potentially be the most impactful \citep{de2020case, brennan2009evaluating}.
% Intuitively, the hope is that an algorithm or LLM can observe minute factors that a human may not pick up on, reduce detectable human bias, or provide further confidence in a human's own decisions.

Most LLM UQ methods have failed to address and evaluate on tasks with aleatoric uncertainty.
% This is mostly due to a focus on MCQA datasets with single, correct answers.
% To address and measure the ability of LLMs to capture aleatoric uncertainty, we need benchmark tasks which truly contain aleatoric uncertainty.
The most popular datasets that LLM UQ papers are evaluated on are QA datasets.
For example, benchmarks such GSM8k \citep{gsm8k}, MMLU \citep{mmlu}, and TriviaQA \citep{triviaQA} all collect questions with a single \emph{correct} answer, labeled and checked by humans.
The existence of a single gold label essentially eliminates the possibility of estimating aleatoric uncertainty, which relies on outcomes themselves not always having the same, correct answer.\footnote{The only remaining possible source of aleatoric uncertainty is \emph{unintentional} noise from annotator mistakes, and that is often removed in followup works seeking to improve a benchmark, e.g. \citet{rucker2023cleanconll}.}

In fact, of the 22 benchmarks used in LLM UQ papers that we evaluated, we found that only one of the 22 benchmarks we analyze contained aleatoric uncertainty \emph{intentionally} \citep[AmbigQA]{min-etal-2020-ambigqa}. 
% We highlight two important points related to aleatoric uncertainty.
% First, the ability of an LLM UQ method to estimate epistemic uncertainty well may not necessarily translate to estimating aleatoric uncertainty \citep{vazhentsev2023hybrid}.
% Second, \citet{baan2022stop} demonstrate that measuring and correcting for calibration on tasks with aleatoric uncertainty requires care, especially in multiclass settings.
A lack of evaluation on benchmarks containing aleatoric uncertainty is not due to the fault of benchmark creators.
Indeed, there exist a plethora of benchmarks designed with aleatoric uncertainty in mind.
For example, SST-5 (with the full five classes, \citet{socher2013recursive}), AmazonReviews \citep{hou2024bridging}, toxic text classification datasets such as ToxiGen \citep{ToxiGen}, and opinion aggregation datasets like ChaosNLI \citep{chaosNLI} contain multiple label annotations per instance with label variation, thus intrinsically encoding the aleatoric uncertainty in the underlying population.

One recent dataset proposed specifically to evaluate LLM UQ methods under aleatoric uncertainty is FolkTexts \citep{folktexts}, which asks an LLM to answer questions about an individual based on open US Census data.
% \vsmargincomment{Given there are these two datasets which have aleatoric uncertainty, can one argue they are good enough benchmarks for this purpose? What else would you want in a benchmark, or would you just want more papers using these two benchmarks?}
Since the dataset has hundreds of thousands of datapoints for each task, given a particular task and set of features, the dataset often contains multiple, possibly conflicting outcomes.
A \emph{calibrated} LLM, therefore, would output reasonable probabilistic estimates on every individual.
Importantly, \citep{folktexts} finds that not all LLM uncertainty elicitation and calibration methods on LLMs perform well under aleatoric uncertainty, further highlighting the need to test for it explicitly.
Further, autoregressive LLMs have been shown to exhibit inconsistent label prediction and probability distributions in such multi-label scenarios~\citep{ma:llmmlc}.

\textbf{Ambiguity and Selective Prediction.} Areas such as ambiguous question answering \citep{cole2023selectively, zhang2021situatedqa} and selective prediction \citep{geifman2017selective,chen2023adaptation} also incorporate elements of aleatoric uncertainty in LLMs.
For example, ambiguous QA datasets such as AmbigQA \citep{min-etal-2020-ambigqa} and Ambiguous Trivia QA \citep{kuhn2022clam} introduce aleatoric uncertainty through linguistic ambiguity of queries.
However, evaluation is usually tied to identifying clarifying questions that would \emph{disambiguate} the original query into a new query with a single correct answer, rather than demanding that an LLM represent the probabilistic ambiguity of the original query.
Additional work in selective prediction also highlights the need for more LLM UQ methods to be tested and verified against aleatoric uncertainty.
For example, \citet{vazhentsev2023hybrid} demonstrate that UQ methods which are designed and tested on only capturing \emph{epistemic uncertainty} can under-perform on selective prediction tasks which contain aleatoric uncertainty.
% \sdmargincomment{This paper sort of justifies vatsals margin comment above, but not sure how to tie the discussion together.}
% They propose a hybrid approach via explicitly considering both epistemic and aleatoric uncertainty to achieve improvements.

These related developments underscore an important point: if LLM UQ methods are only ever evaluated against epistemic uncertainty --- and not tested on their ability to quantify and express irreducible, data-inherent uncertainty present in many real-world scenarios --- then UQ methods may end up less useful for user-centric downstream applications like selective prediction or ambiguous QA.
This blind spot reinforces the need to test against aleatoric uncertainty as a central tenet.

\subsection{Distribution Shift}
\label{sec:distribution-shift}
Real-world LLM users are diverse in many ways, including geographically \citep{wildchat, dunn2024pre}, 
 and culturally \citep{gershcovich2024challenges}, with tangible impacts on downstream applications. 
For example, \citet{xing2022age} demonstrate that the way that users interact with conversational search agents differs in complexity and length by user age.
Robustness to these variations is often not measurable when evaluating a LLM UQ method on a train / test split of a single benchmark.
Instead, robustness to query variations requires that LLM UQ methods remain useful in the face of a third kind of uncertainty: \emph{distribution shift}.

Many LLM UQ methods are focused on \emph{post-processing} the outputs of the model in a way which improves calibration or uncertainty estimation.
Such methods usually fall within the regime of supervised \emph{parameter learning}, i.e., the method requires learning some parameters from a set of data.
For example, the well-known Temperature or Platt scaling UQ methods \citep{guo2017calibration, platt1999probabilistic} ---- variants of which are often applied in LLM UQ \citep{ulmer2024calibrating, xie2024calibrating} --- require hold-out data in order to fit re-scaling parameters for model logits or probabilistic predictions.
Even unsupervised \emph{parameter-free} LLM UQ methods such as verbalized confidence scoring \citep{tian2023just, xiong2023can} often apply post-hoc calibration to improve the final uncertainty estimate on a particular task \citep{wang2024calibrating}.
For such methods, the exact hold-out data used makes a difference in the learned parameters, and hence, the performance, of the UQ method. 

It is therefore vital to assess performance on data which is \emph{out-of-distribution} (OOD): i.e., queries which \emph{look different} from the training data used to learn the original parameters of the LLM UQ method. 
Consistent failure of LLM UQ methods to recognize or adapt to distribution shifts may potentially undermine \emph{user trust} in the LLM's reliability, which can result in decreased utility of the joint human-AI system \citep{srinivasan2025adjust}.

Unfortunately, we find that robustness to distribution shift was discussed and tested in only 10 out of the 19 supervised, parameter-learning LLM UQ methods papers we evaluated.
We believe that testing post-hoc calibration methods under distribution shift should be a requirement of any LLM UQ paper. 
Evaluating under distribution shift as a standard practice would benefit LLM UQ methods in real-world settings where user-queries may change over time or shift with new demographics, and also when LLM UQ methods are used in downstream tasks --- such as hallucination detection \citep{detommaso2024multicalibration, zhang2023enhancing} or selective prediction \citep{chen2023adaptation} --- where OOD performance is essential for reliability.

\subsection{Recommendations}
To address the concerns raised in this section, we provide two simple and actionable recommendations: 

\begin{enumerate}[label=\textbf{R\arabic*.}, leftmargin=*]
    \item[\textbf{R4.}] Evaluate methods on datasets with intrinsic, intentional aleatoric uncertainty (like FolkTexts \citep{folktexts} or ChaosNLI \citep{chaosNLI}).
    
    \item[\textbf{R5.}] For works proposing supervised LLM UQ methods, ensure that you are testing on settings with distribution shift. Evaluating under distribution shift is especially important for methods utilizing conformal prediction guarantees \citep{cherian2024large, quach2024conformal}, since theoretical conformal guarantees usually do not hold under distribution shift \citep{aolaritei2025conformal}.
\end{enumerate}

\section{Metrics and Underscoring the Need for Human Studies}\label{sec:hci}

We argue that metrics for optimizing and evaluating LLM UQ methods may not sufficiently capture the ability of LLMs to foster effective and meaningful human-AI collaboration.
In particular, most papers we surveyed focus on standard calibration measures such as Brier score, ECE, or AURUC on simple QA tasks.\footnote{In \Cref{appx:smECE}, we also argue a switch from ECE to \emph{smECE} \citep{blasiok2024smooth}.}
We argue for a focus on metrics that correlate with \emph{human uplift on real-world decision tasks}.
This user-centric shift requires advancement in both the theory of human-AI collaboration and practical human studies on the utility of algorithmic assistance with quantified uncertainty.
Although some studies have already been conducted in \emph{classical UQ} settings \citep{corvelo2025human, alur2023auditing}, the interplay between generated language, UQ, and usefulness to users has been drastically under-considered. 
Of the issues and directions that we raise in each section, this issue requires the most thought, focus, and careful research to execute.

\subsection{Inspiration from Classical UQ: Calibration Metrics for Real-world Usefulness}
\label{sec:real-world-metrics}

When an LLM assists a doctor in diagnosing a complex case or a software engineer in evaluating potential bugs in a section of code, the fundamental goal of UQ extends beyond achieving good ECE or Brier score on the task. 
% Instead, the crucial test is 'human uplift': does the articulated uncertainty inform individuals to make demonstrably better or more efficient decisions?
Instead, a common desiderata of the LLM UQ literature is to make language models more \emph{useful and transparent} as a human aide \citep{yang2024verbalized, tian2023just, sung2025grace, li2024graph}.
Computing the uncertainty of a model on a particular task or data point is one facet of this goal.
However, as the aforementioned examples demonstrate, simply computing a calibrated probability is often not the end of the story.
In order to declare success, LLM UQ --- like many other research areas related to modern LLM applications \citep{kapoor2025normalAI} --- requires \emph{human uplift studies}.
In particular, do humans with access to algorithmic advice with quantified uncertainty perform better on real-world tasks than humans with (i) algorithmic advice but no quantified uncertainty; and (ii) no algorithmic advice at all?

Both the theory and the human-computer interaction (HCI) communities have explored aspects of how uncertainty should be presented to users in order to \emph{provide uplift}; i.e., help users improve their decision making by accessing algorithmic advice.
In classical (non-LLM) UQ, results are still emerging, but they already point to a somewhat surprising trend: simply calibrating the probabilities does not seem to be the best way to present uncertainties to users.
For example, \citet{donahue2022human} prove a necessary and sufficient condition on the \emph{complementariness} of human and algorithmic predictions in order for uplift to occur in a simple theoretical framework.
Importantly, model calibration (or accuracy) \emph{does not} always satisfy the required condition. 

There is a host of additional work in UQ which echoes similar results: \citet{vodrahalli2022uncalibrated} show that humans on a real task perform better when given access to algorithmic advice which is \emph{uncalibrated} rather than calibrated.
\citet{cao2024designing} also show that model calibration is not a sufficient condition for uplift in human studies, and focus on designing uncertainty presentation schemes which are more effective.
Finally, \citet{corvelo2023human} prove that calibrated models will \emph{always be suboptimal} when aggregating human and algorithmic predictions.
Instead, they propose obtaining models which are not necessarily calibrated overall, but instead are \emph{human-aligned} in a formal sense defined by how human and algorithmic predictions relate to one another. 
% \emph{calibrated on the decision level sets of humans} via a notion called \emph{multicalibration}.
They then demonstrate that \emph{human-alignment} empirically correlates with improved algorithmic assisted human performance in human studies ran on a synthetic task \citep{corvelo2025human}. 
Lastly, \citet{vasconcelos2025generation} show that LLM UQ methods may need to be further modified to provide uplift for programmers on real-world coding tasks.

Together, these works reveal a striking disconnect between research on human-AI collaboration and LLM UQ.
At face value, LLM UQ research claims to be focused on the goal of making language models more reliable and trustworthy \citep{li2024graph}.
At a deeper level, however, there exists very little research which has \emph{directly} confirmed whether LLM UQ is actually useful to humans on real decision making tasks.
In particular, human-uplift studies from both the HCI and theory communities seem to demonstrate that calibrated probabilities --- the gold standard of LLM UQ --- may not be the correct metric for optimal human-AI collaboration.
Nonetheless, many LLM UQ papers remained as focused as ever on providing simple, calibrated probabilities on tasks like factual question answering or knowledge retrieval.
% The trend from both theoretical algorithmic prediction frameworks and real human studies does not seem to have trickled down to the pure UQ literature: 
While optimizing LLM UQ metrics like ECE and Brier score may already be useful for applications like selective prediction \citep{li2023coannotating} or pure capabilities research like reasoning \citep{taubenfeld2025confidence}, optimizing these same metrics not necessarily imply that we will get better collaborative human-LLM performance ``for free'' out of the box.
% it represents a disconnect with the HCI research community which would potentially stand to benefit from any improvements proposed by novel UQ methods.

\textbf{New Uncertainty Presentation Schemes for LLMs.} 
As articulated in \citet[Section 3.5]{liao2023ai}, in addition to simply presenting probabilities or confidences to users, LLM UQ techniques can look very different than classical UQ ones through taking specific advantage of the medium of language.
For example, LLM UQ methods may utilize unique ways of introducing vaguer claims in order to refrain from complete abstention when highly uncertain \citep{jiang2025conformal}, or use linguistic hedging~\citep{mielke2022reducing} and anthropomorphic language~\citep{kim2024m} to convey uncertainty in non-numeric ways to users.
Such methods are unique to LLM UQ and should be further investigated to understand how they impact human uplift on decision tasks.

\subsection{Recommendations}

To ameliorate the aforementioned shortcomings of current LLM UQ evaluation metrics, we advocate for three research directions, which we outline below.
Via these proposed approaches, we push for \emph{better cohesion} between NLP researchers developing LLM UQ methods and the HCI community.
Ideally, this will create and accelerate feedback loops between both communities, allowing for more rapid iteration and investigation into LLM UQ approaches which provide measurable human uplift.
% \sdmargincomment{I'm open to combining this into only two recommendations if you can figure out how.}

\begin{enumerate}[label=\textbf{R\arabic*.}, leftmargin=*]
    \item[\textbf{R6.}] Adapt findings from existing studies in classical UQ on human-aligned calibration~\citep{corvelo2025human, vodrahalli2022uncalibrated}  to uncertainty-augmented LLM advice.

    \item[\textbf{R7.}] Develop studies which attempt to understand the relationship between LLM UQ metrics (like ECE, Brier score) and human-AI team performance.

    \item[\textbf{R8.}] Incentivize more studies on how the presentation of LLM uncertainty impacts user reliance.  
\end{enumerate}

Recommendation \textbf{R6} is fairly straightforward and builds upon the numerous UQ papers with human studies mentioned in \Cref{sec:real-world-metrics}.
It is possible that presenting probabilities of generated text via LLM UQ methods may have a differing effect on users when compared with the classical UQ setting.
Language responses may offer explanations and anthropomorphic cues, whereas predictive classical models used in prior human studies do not.
Understanding the effect of presenting uncertainty estimates alongside LLM-generated responses on human-AI collaboration outcomes is vital.

Our next recommendation \textbf{R7} is more fundamental: it articulates a necessity to better understand the metrics that LLM UQ seeks to optimize.
The community stands to benefit from comprehensive analyses on whether metrics like ECE or Brier score \emph{correlate} with increased downstream performance by users.
Understanding this relationship between metrics and downstream utility could inform the LLM UQ community on whether to continue optimizing for metrics such as ECE, or potentially switch to other measurable metrics identified as important by human studies.
As an example, the seminal natural language summarization metric ROUGE was shown to correlate with human summary judgments, making researchers more confident that it is a reasonable metric to hillclimb on \citep{lin-2004-rouge}.
Furthermore, other metrics like BLEU in machine translation were shown to have the opposite behavior --- that is, they were shown to at times correlate poorly with human judgments \citep{callison2006re} --- shifting attention to the development of new metrics \citep{freitag2022results}.
Without evidence that calibration metrics are correlated with utility to users, our community risks "flying blind" by choosing ad hoc evaluation metrics.

Lastly, \textbf{R8} points to the need to better understand the effect of LLM uncertainty presentation on user reliance.
It is important for researchers to understand what types of LLM UQ uncertainty presentation schemes --- e.g., displaying claim-level numeric probabilities \citep{detommaso2024multicalibration}, hedged linguistic uncertainty \citep{band2024linguistic}, hedges with anthropomorphic cues~\citep{kim2024m} --- perform best in different human decision contexts.
Such research would directly steer further research in the LLM UQ community, and as such, should be encouraged to be published in ML / NLP conferences, as opposed to being restricted to only HCI venues such as CHI.

% Evidence:

\section{Alternative Views and Conclusions}
\label{sec:alternative-views}

We present two potential alternative views that a researcher in LLM UQ may make regarding our call for a re-focusing of the subfield on human-centric utility.

\emph{Current LLM UQ research is laying essential groundwork with technical metrics and simpler benchmarks. Complex evaluations can come later.}

% Alternative views

\textbf{Rebuttal.} Relying only on particular metrics or simplistic benchmarks in the nascent stages of a research subfield potentially risks establishing a trajectory which may not exactly align with the ultimate goal of human-centric utility in down-stream decision making.
As an example of this in action, we lean on the developments over the past decade in explainable-AI (XAI).
% We believe that the developments in LLM UQ will mirror those in the explainable-AI (XAI) subfield.
In XAI, what started primarily as a technical and quantitative research area transformed into one with human-AI interaction at its core.
This perspective change occurred only once a host of work realized that i) always providing model explanations may sometimes \emph{reduce} the performance of the AI-assisted human \citep{paleja2021utility, buccinca2020proxy, alufaisan2021does, liao2022connecting, joshi2023are}; and 2) learning when and how to jointly model AI assistance and human behavior cannot always be solved with purely technical solutions, and is highly context dependent \citep{rong2023towards}.
Indeed, LLM UQ can be considered as one avenue of research \emph{within} the broader field of XAI, and as such, we should apply many of the same insights from XAI here.

\emph{Current benchmarks are scalable and allow rapid iteration. Your proposed evaluations are too resource-intensive and would slow innovation and evaluation.}

\textbf{Rebuttal.} 
While the cost of evaluating current LLM UQ benchmarks is certainly cheaper than running human studies, the \emph{long-term} cost of deploying UQ solutions which erode user trust or lead to lower decision system utility is arguably far greater.
Over-zealous and rushed deployments of new technologies have historically had lasting impact on user trust.
For example, in 2021, \citet{wong2021external} showed that healthcare software company Epic's sepsis classification tool performed poorly at deployment time (in part due to \emph{distribution shift} similar to that described in \Cref{sec:distribution-shift}).
In addition, because of how the system was implemented, clinicians and nurses ended up having such low trust in the system that they tried to minimize its involvement in decisions \citep{gichoya2023ai, habib2021epic}.
Such examples underscore the need for human-centric technology development in order to yield effective and efficient human-uplift on important real-world tasks \citep{shneiderman2020human}.

% \subsection{Conclusion}
\textbf{Conclusion.} In this work, we have laid out what we believe are foundational issues in the evaluation of LLM UQ methods. 
In particular, we highlight three problematic practices:  1) evaluating on benchmarks with low ecological validity; 2) only considering epistemic uncertainty; and 3) hill-climbing on metrics that are not necessarily indicative of downstream utility. 
We further provide concrete recommendations for overcoming each of these issues, either through revised practices or new research directions. 
Although numerous, we believe these problems can be addressed by the many researchers pursuing this subfield through the adoption of a more human-centered perspective.

\newpage

\bibliographystyle{apalike}
\bibliography{refs}

\appendix
\newpage

\section{ECE and Measuring Calibration in Practice}
\label{appx:smECE}
There may be other reasons to shy away from solely optimizing ECE when measuring calibration error for LLMs.
A recent sequence of related work has investigated the fundamental problem of \emph{measuring} calibration error \citep{blasiok2023unifying, blasiok2024smooth, chidambaram2025reassessing}.
At a high level, these works together demonstrate that ECE --- the most popular calibration error notion --- can be highly unstable and not-robust.
This is due to a discretization / binning procedure that is necessary when computing ECE because of measurability issues with real-valued probabilistic predictions.
\citet{blasiok2024smooth} instead propose a related but robust notion termed SmoothECE (smECE), which gets around the binning requirement by applying kernel smoothing.
\cite{hansen2024when} further demonstrate that using smECE rather than ECE can change conclusions about what model is better calibrated on a particular task, especially within the small sample regime (<5k datapoints).

ECE, however, remains a popular metric for measuring calibration error of LLM UQ methods.
We found that 16 papers used ECE to measure calibration error, and further, many of these papers had evaluation sets which were smaller than 5k datapoints.
For example, both \citet{li2024graph} and \citet{yang2024bayes} used ECE to measure performance on evaluation sets of around 2k datapoints.
At this scale of dataset size, smECE should be preferred.
Among all papers we analyzed, the only paper which utilized smECE to measure calibration error was \cite{ulmer2024calibrating}.
Indeed, \citet[Table 3]{ulmer2024calibrating} demonstrates that measuring calibration error with smECE vs. ECE can result in a different ordering in the quality of LLM UQ methods applied to Vicuna and GPT-3.5. 
Due to the difficulty and cost of obtaining LLM predictions at larger dataset scales, we believe that the community stands to benefit from utilizing smECE as a drop-in replacement for ECE.
Doing so may help ensure that any technical advancements in calibration error remain robust and statistically significant even with test dataset size restrictions. 

\section{LLM UQ Methods Annotation}
\label{sec:methods-annotated}
\newcommand{\paperannotation}[7]{%
  \item \textbf{#1}
  % \subsection{#1}
  \begin{enumerate}
    \item Type of uncertainty quantification (short / long form generation, conformal, etc.).
    \begin{itemize}
      \item #2
    \end{itemize}

    \item Datasets used to evaluate proposed LLM UQ method.
    \begin{itemize}
      \item #3
    \end{itemize}

    \item Test performance under distribution shift? (yes / no). Can also be N/A for methods which are unsupervised.
    \begin{itemize}
      \item #4
    \end{itemize}

    \item Supervised, semi-supervised, or unsupervised method?
    \begin{itemize}
      \item #5
    \end{itemize}

    \item Metrics used to evaluate performance.
    \begin{itemize}
      \item #6
    \end{itemize}

    \item Human uplift study?
    \begin{itemize}
      \item #7
    \end{itemize}
  \end{enumerate}
}

In this section, we provide our annotation for each LLM UQ method paper which we include in our survey.
\begin{enumerate}[label=P\arabic*., leftmargin=*]

\paperannotation{Can LLMs Express Their Uncertainty? An Empirical Evaluation of Confidence Elicitation in LLMs \citep{xiong2023can}}
{Verbalized confidence via prompting, sampling, and aggregation techniques.}
{SportUND (commonsense),
StrategyQA (commonsense),
GSM8K,
SVAMP,
Date Understanding (symbolic reasoning),
Object Counting (symbolic reasoning),
MMLU-Law,
MMLU-BizEthics}
{N/A}
{Unsupervised, but has hyperparameters which can influence the outcome of the method.}
{ECE, AUROC, AUPRC}
{No}

\paperannotation{Generating with Confidence: Uncertainty Quantification for
Black-box Large Language Models \citep{lin2024generating}}
{A simple measure for the ``semantic dispersion'' of a set of generated short-form responses.
Can be a reliable predictor of the quality of LLM responses.}
{CoQA,
TriviaQA,
Natural Questions}
{N/A}
{Unsupervised.}
{AUROC, AUARC}
{No}

\paperannotation{Semantic Uncertainty: Linguistic Invariances for Uncertainty Estimation in Natural Language Generation \citep{kuhn2023semantic}}
{Short form UQ via clustering multiple generations and calculating entropy.}
{CoQA, TriviaQA}
{N/A}
{Unsupervised}
{AUROC}
{No}

\paperannotation{Shifting Attention to Relevance: Towards the Predictive Uncertainty
Quantification of Free-Form Large Language Models \citep{duan2023shifting}}
{Short form UQ via multiple generations and re-weighing at the token / sentence level.}
{CoQA, TriviaQA, SciQ, MedQA, MedMCQA}
{N/A}
{Unsupervised}
{AUROC}
{No}

\paperannotation{Kernel Language Entropy: Fine-grained Uncertainty Quantification for LLMs from Semantic Similarities \citep{nikitin2024kernel}}
{Short form UQ via clustering multiple generations and calculating entropy in a more refined approach than semantic entropy (is a generalization of that work)}
{TriviaQA, SQuAD, BioASQ, NQ, SVAMP}
{N/A}
{Unsupervised}
{AUROC, AUARC}
{No}

\paperannotation{Mars: Meaning-aware response scoring for uncertainty estimation in generative llms \citep{bakman2024mars}}
{Unsupervised UQ for short QA responses.}
{TriviaQA, 
Natural Questions,
WebQA}
{N/A}
{Unsupervised}
{AUROC}
{No}

\paperannotation{Uncertainty Estimation in Autoregressive Structured Prediction \citep{malinin2021uncertainty}}
{Unsupervised UQ for short QA responses.}
{newstest14 and LibriSpeech (translation datasets)}
{Yes (out of pre-training domain)}
{Unsupervised}
{NLL, AUPR, PRR}
{No}

\paperannotation{Quantifying Uncertainty in Answers from Any Language Model and Enhancing Their Trustworthiness \citep{chen2023quantifying}}
{Unsupervised UQ for short QA responses.}
{GSM8k, SVAMP, CSQA, TriviaQA}
{N/A}
{Unsupervised}
{AUROC}
{No}

\paperannotation{Enhancing Uncertainty-Based Hallucination Detection with Stronger Focus \citep{zhang2023enhancing}}
{Unsupervised, reference free UQ method for hallucination detection at the sentence level.}
{WikiBio}
{N/A}
{Unsupervised}
{AUC-PR}
{No}

\paperannotation{Language Models (Mostly) Know What They Know \citep{kadavath2022language}}
{P(True) token probabilities in short form UQ, also UQ fine-tuning.}
{TriviaQA, Lambada, Arithmetic, GSM8k, and Codex HumanEval}
{Yes}
{Supervised and Unsupervised}
{ECE}
{No}

\paperannotation{SELFCHECKGPT: Zero-Resource Black-Box Hallucination Detection
for Generative Large Language Models \citep{manakul2023selfcheck}}
{Detect hallucinations in long form generated text via UQ methods.}
{WikiBio}
{N/A}
{Unsupervised}
{AUC-PR}
{No}

\paperannotation{Graph-based confidence calibration for large language models \citep{li2024graph}}
{Confidence of short form generation.}
{CoQA, TriviaQA}
{Yes}
{Supervised}
{ECE, AUROC, Brier}
{No}

\paperannotation{Calibrating Large Language Models Using Their Generations Only \citep{ulmer2024calibrating}}
{Train auxilary model to predict confidence of short form generations of LLM}
{TriviaQA }
{No}
{Supervised}
{Brier, ECE, smECE, AUROC}
{No}

\paperannotation{Bayesian Low-Rank Adaptation for Large Language Models \citep{yang2024bayes}}
{Train LoRA Bayesian ensembles fine-tuned on target task.}
{MC / common sense reasoning tasks: WG-S, ARC-C, ARC-E, OBQA, WG-M, BoolQ, MMLU}
{Yes}
{Supervised}
{Acc, ECE, NLL}
{No}

\paperannotation{Reconfidencing LLM Uncertainty from the Grouping Loss Perspective \citep{chen2024reconfidencing}}
{Short form uncertainty quantification on answers}
{Propose new dataset, short answer but derived for fairness purposes (has group information)}
{No}
{Supervised}
{Brier, CL, GL}
{No}

\paperannotation{Reducing conversational agents’ overconfidence
through linguistic calibration \citep{mielke2022reducing}}
{Train a "calibrator" that tries to predict the correctness of the LLM based on its internal state.}
{TriviaQA}
{No}
{Supervised}
{ECE, Max CE, NLL}
{No}

\paperannotation{Multicalibration for Confidence Scoring in LLMs \citep{detommaso2024multicalibration}}
{Supervised UQ for short QA respones}
{BigBench, MMLU, triviaQA,
OpenBookQA, TruthfulQA, and MathQA}
{No}
{Supervised}
{Group average squared calibration error (gASCE), ASCE (similar to ECE).}
{No}

\paperannotation{Teaching Models to Express Their Uncertainty in Words \citep{lin2022teaching}}
{Prompts LLM to generate confidence as a sequence of tokens}
{CalibratedMath}
{Yes}
{Supervised}
{Brier, ECE}
{No}

% \paperannotation{\url{https://aclanthology.org/2023.acl-long.652.pdf}}
% {N/A}
% {Five toxicity datasets
% Three multi-class tasks with high ambiguity: 20NewsGroups, SST-5, AmazonReviews}
% {Kind of yes}
% {Supervised}
% {AUC-RC}
% {N/A}

% \paperannotation{\url{https://arxiv.org/pdf/2209.15558}}
% {N/A}
% {Summarization and Translation datasets. Too many to list, but should have both epistemic and aleatoric, since they discuss this.}
% {Yes}
% {Supervised}
% {AUROC}
% {N/A}

% \paperannotation{\url{https://arxiv.org/pdf/2504.14154}}
% {N/A}
% {MMLU, MedCQA, TriviaQA, CoQA}
% {Yes}
% {Supervised}
% {N/A}
% {N/A}

% \paperannotation{\url{https://arxiv.org/pdf/2406.09714}}
% {Filters output to remove incorrect claims in the generation process}
% {MedLFQA, Wikipedia Bios}
% {No (They acknowledge that their method may not work well o.o.d.)}
% {Supervised}
% {N/A}
% {N/A}

\paperannotation{Do Not Design, Learn: A Trainable Scoring Function
for Uncertainty Estimation in Generative LLMs \citep{yaldiz2024not}}
{Trainable / learnable response scoring function for LLMs. Follow-up work on \citep{bakman2024mars}.}
{TriviaQA
NaturalQA
WebQA
GSM8K}
{Yes}
{Supervised}
{AUROC, PRR}
{No}

\paperannotation{Conformal Linguistic Calibration: Trading-off between Factuality and Specificity \citep{jiang2025conformal}}
{Modify the output to make language used more vague to better represent the uncertainty of the model.}
{SimpleQA, Natural Questions}
{No}
{Supervised}
{Conformal coverage / factuality requirement.}
{No}

\paperannotation{Calibrating Expressions of Certainty \citep{wang2024calibrating}}
{Calibrating linguistic expressions of certainty, e.g., “Maybe” and “Likely”, in both humans and language models.}
{SciQ, TruthfulQA}
{No}
{Supervised}
{ECE and Brier Score.}
{Test methods on humans.}

\paperannotation{QA-Calibration of Language Model Confidence Scores \citep{manggalaqa}}
{Group-wise calibration guarantees for QA}
{MMLU, TriviaQA, SciQ, BigBench, OpenBookQA}
{No}
{Supervised}
{AUAC, calibration error}
{No}

\paperannotation{Unconditional Truthfulness: Learning Conditional Dependency for Uncertainty Quantification of Large Language Models \citep{vazhentsev2024unconditional}}
{Learn conditional dependency between generation steps in order to estimate uncertainty.}
{Text generation: CNN/DailyMail, XSUM, SamSum
Long-form QA: MedQUAD, PubMedQA, TruthfulQA
Short-form QA: SciQ, CoQA, TriviaQA}
{No}
{Supervised}
{Prediction Rejection Ratio (PRR, selective prediction metric).}
{No}

% \paperannotation{Learning Confidence for Transformer-based Neural Machine Translation}
% {N/A}
% {Machine Translation datasets.}
% {yes}
% {Supervised}
% {N/A}
% {N/A}

\paperannotation{Just Ask for Calibration: Strategies for Eliciting Calibrated Confidence
Scores from Language Models Fine-Tuned with Human Feedback \citep{tian2023just}}
{Unsupervised prompting methods, and also supervised method which uses a calibration hold-out set to choose confidence levels.}
{TriviaQA, SciQA, TruthfulQA}
{No}
{Mix of supervised / unsupervised}
{ECE, ECE-t, BS-t, AUC}
{No}

\paperannotation{Do LLMs Estimate Uncertainty Well in Instruction Following? \citep{heo2024llms}}
{Systematic evaluation of LLM UQ in instruction following.}
{IFEval}
{N/A}
{Unsupervised}
{AUROC}
{No}

\paperannotation{Improving Uncertainty Estimation through Semantically Diverse Language Generation \citep{aichberger2025improving}}
{Estimate aleatoric semantic uncertainty via steering multiple generations}
{TruthfulQA, CoQA, TriviaQA}
{N/A}
{Unsupervised}
{Rouge-L, Rouge-1, BLEURT, AUROC}
{No}

\paperannotation{Can Large Language Models Faithfully Express Their Intrinsic Uncertainty in Words? \citep{yona2024can}}
{Evaluate LLM UQ with a new metric called faithfulness. Argue that LLMs ignore intrinsic uncertainty. Also test a few-shot prompting-based LLM UQ method Uncertainty+.}
{Natural Questions, PopQA}
{N/A}
{N/A}
{Faithfulness}
{N/A}

\paperannotation{Confidence Under the Hood: An Investigation into the Confidence-Probability Alignment in Large Language Models \citep{kumar2024confidence}}
{Connect confidence stated by LLM (conveyed in model response) to the confidence quantified by token probabilities.}
{CommonsenseQA, QASC, RiddleQA, OpenBookQA, ARC}
{N/A}
{Unsupervised}
{Spearman correlation}
{No}

% \paperannotation{\url{https://arxiv.org/pdf/2502.10709}}
% {N/A}
% {N/A}
% {N/A}
% {N/A}
% {N/A}
% {N/A}

\paperannotation{Language Model Uncertainty Quantification with Attention Chain \citep{li2025language}}
{Uses attention weights to find semantically critical tokens in a backtracking procedure. Allows estimation of the marginal probabilities of answer tokens.}
{GSM8K, MATH, BigBench}
{No}
{Unsupervised}
{AUROC, ECE}
{No}

\paperannotation{Preserving Pre-trained Features Helps Calibrate Fine-tuned Language Models \citep{he2023preserving}}
{Calibrating fine-tuned language models results in poor performance under distribution shift. Propose method to preserve pre-trained features in order to improve calibration.}
{WikiText-103, SWAG, HellaSWAG, MNLI, SNLI}
{Yes}
{Supervised}
{ECE + reliability diagram}
{No}

% \paperannotation{Approaching Human-Level Forecasting with Language Models \citep{halawi2024approaching}}
% {N/A}
% {Comparing against crowd-sourced forecasting sites}
% {N/A}
% {N/A}
% {Brier score, accuracy}
% {N/A}

\paperannotation{Uncertainty Estimation and Quantification for LLMs: A Simple
Supervised Approach \citep{liu2024uncertainty}}
{Use hidden layer features for UQ in a supervised manner.}
{CoQA, TriviaQA, MMLU, WMT14 Machine Translation}
{Yes}
{Supervised}
{AUROC}
{No}

\paperannotation{Just rephrase it! Uncertainty estimation in closed-source language models via multiple rephrased queries \citep{yang2024just}}
{Rephrase the prompt to sample multiple outputs, aggregate to get confidence.}
{ARC, openbook QA, MMLU}
{N/A}
{Unsupervised}
{Acc, ECE, TACE, Brier, AUROC}
{No}

\paperannotation{Uncertainty Quantification with Pre-trained Language Models: A Large-Scale Empirical Analysis \citep{xiao2022uncertainty}}
{Benchmark various LLM UQ methods on NLP tasks.}
{Sentiment analysis: IMDB, Yelp reviews. NLI: MNLI, SNLI. Commonsense reasoning: SWAG, HellaSWAG}
{Yes}
{Supervised}
{ECE}
{No}

\paperannotation{Beyond Confidence: Reliable Models Should Also Consider Atypicality \citep{yuksekgonul2023beyond}}
{Estimate the \emph{atypicality} of inputs in order to improve LLM UQ.}
{IMDB, TREC, AG News}
{Yes}
{Supervised}
{ECE}
{No}

\paperannotation{Overconfidence is Key: Verbalized Uncertainty Evaluation in Large Language and Vision-Language Models \citep{groot2024overconfidence}}
{Prompts LLM to output confidence intervals}
{SST (sentiment analysis)
GSM8k (math)
CoNLL (NER)
Japanese Uncertain Images (for VLMs only)}
{N/A}
{Unsupervised}
{ECE, MaxCE, NCE}
{No}

\paperannotation{Multi-group Uncertainty Quantification for Long-form Text Generation \citep{liu2024multi}}
{Uncertainty quantification for factual correctness in long form text generation}
{BIO-NQ (subset of natural questions corresponding to biographies)}
{No}
{Supervised}
{ASCE (average squared calibration error), max gASCE, average gASCE, Brier score}
{No}

\paperannotation{LACIE: Listener-Aware Finetuning for Confidence Calibration in Large Language Models \citep{stengel2024lacie}}
{Cast calibration as preference optimization problem with a speaker and listener.}
{TriviaQA
TruthfulQA}
{Yes}
{Supervised}
{ECE, AUROC, Precision, Recall of user}
{Yes, human evaluation with annotators accepting or rejecting LLM answers.}

\end{enumerate}

\end{document}